\documentclass[pdflatex,sn-mathphys-num]{sn-jnl}%

\usepackage{graphicx}%
\usepackage{multirow}%
\usepackage{amsmath,amssymb,amsfonts}%
\usepackage{amsthm}%
\usepackage[title]{appendix}%
\usepackage{xcolor}%
\usepackage{textcomp}%
\usepackage{manyfoot}%
\usepackage{booktabs}%
\usepackage{algorithm}%
\usepackage{algorithmicx}%
\usepackage{algpseudocode}%
\usepackage{listings}%
\usepackage{arydshln}%
\usepackage{array}%
\usepackage[T1]{fontenc}
\usepackage{lmodern}
\usepackage{fix-cm}

\begin{document}

\title[Article Title]{Instruction-Following Evaluation of Large
Vision-Language Models}

\author[1]{\fnm{Daiki} \sur{Shiono}}\email{daiki.shiono.s1@dc.tohoku.ac.jp}

\author[1]{\fnm{Shumpei} \sur{Miyawaki}}\email{shumpei.miyawaki.b7@tohoku.ac.jp}

\author[1,2]{\fnm{Ryota} \sur{Tanaka}}\email{ryota.tanaka@ntt.com}

\author[3]{\fnm{Jun} \sur{Suzuki}}\email{jun.suzuki@tohoku.ac.jp}

\affil[1]{\orgdiv{Graduate School of Information Sciences},
  \orgname{Tohoku University}, \orgaddress{\street{Aramakiaza Aoba
    6-3-09, Aoba-ku}, \city{Sendai}, \postcode{980-8579}, \state{Miyagi},
\country{Japan}}}

\affil[2]{\orgdiv{NTT Human Informatics Laboratories}, \orgname{NTT
  Corporation}, \orgaddress{\street{1-1
    Hikarinooka}, \city{Yokosuka}, \postcode{239-0847}, \state{Kanagawa},
\country{Japan}}}

\affil[3]{\orgdiv{The Center for Language AI Research},
  \orgname{Tohoku University}, \orgaddress{\street{Kawauchi 41,
Aoba-ku}, \city{Sendai}, \postcode{980-8576}, \state{Miyagi}, \country{Japan}}}

\abstract{
  Following the initial flourishing of large language models (LLMs),
  there has been a surge in proposed large vision-language models
  (LVLMs) that integrate LLMs with vision capabilities.
  However, it has been observed that LVLMs, after tuning to visual
  instruction using commonly used training datasets, often fail to
  exhibit the instruction-following ability that was present in the
  LLM before integration, leading to results in which they do not
  follow task instructions as expected.
  This study quantitatively demonstrates that LVLMs'
  instruction-following ability declines after fine-tuning and
  analyzes its underlying causes.
  In particular, we constructed new training datasets highlighting
  whether the output format is specified. Then, we investigated how
  explicitly indicating the output format during fine-tuning affects
  LVLMs' instruction-following ability.
  Our quantitative evaluation confirmed that LVLMs'
  instruction-following ability declines after fine-tuning with
  commonly used datasets.
  Furthermore, we found that LVLMs trained with datasets, including
  instructions on output format, tend to follow instructions more
  accurately than models that do not.
  These findings suggest that including samples with instructions on
  output format during (visual) instruction tuning may help mitigate
  the decline in instruction-following abilities.
}


\maketitle
\newpage

\section{Introduction}\label{sec:intro}
Large language models (LLMs) have seen remarkable technical progress,
leading to numerous open-source large vision-language models (LVLMs),
such as BLIP-2~\cite{li2023blip}, LLaVA~\cite{liu2023llava} and
Qwen2-VL~\cite{wang2024qwen2}.
These LVLMs take images and texts as input and generate texts as
outputs. LVLMs can leverage the advanced linguistic inference
capabilities of an LLM that has been trained on a diverse and
large-scale language corpus, using the LLM as a text generator.
\begin{figure}[t]
  \small
  \centering
  \includegraphics[width=0.80\columnwidth]{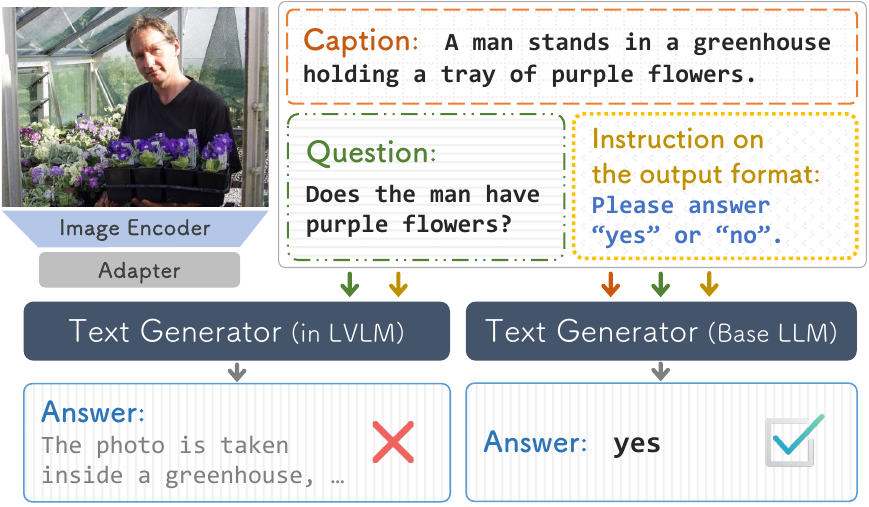}
  \caption{LVLMs (left) show lower instruction-following ability than
    LLMs (right). We examine this gap quantitatively and explore the
  factors that contribute to reductions in instruction-following ability.}
  \label{fig:research_overview}
\end{figure}
LVLMs with a high ability to follow instructions have demonstrated
superior task generalization performance in numerous vision-language
tasks~\cite{wei2021finetuned, liu2023llava}, including answering
visual-based questions~\cite{Antol_2015_ICCV, hudson2019gqa}.
However, as shown in Figure~\ref{fig:research_overview}, while the
LLM, before it is integrated into the LVLM, can follow instructions
and generate correct responses, it has been \textbf{qualitatively}
confirmed by \citet{fu2023mme} that LVLMs, after visual instruction
tuning, could fail to follow instructions, generating unintended answers.

\begin{figure*}[ht]
  \small
  \centering
  \includegraphics[width=\columnwidth]{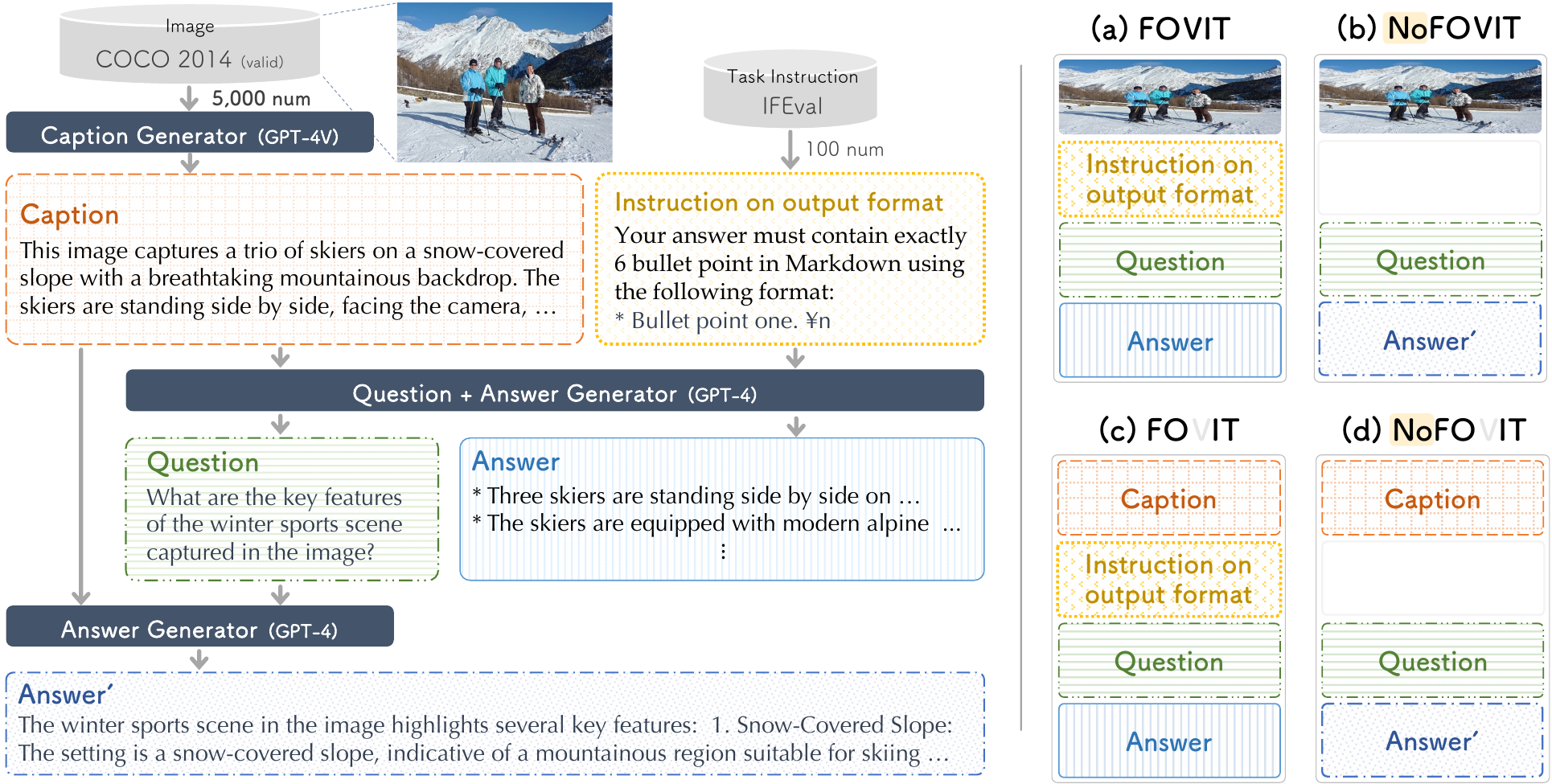}
  \caption{Example of a (visual) instruction training synthetic
    dataset consisting of \texttt{COCO} images, with captions generated
    by GPT-4V, instructions on the output format extracted from the
    \texttt{IFEval} dataset, and question texts and answers generated
  by GPT-4 (Section \ref{subsec:create_instruction_datasets}).}
  \label{fig:data_generation}
\end{figure*}

Therefore, this study \textbf{quantitatively} evaluates LVLMs'
ability to follow instructions, revealing, for the first time, a
decline in this ability.
We also specify factors behind the diminished ability of LVLM to
follow instructions (Figure~\ref{fig:research_overview}).
First, we create new datasets for fine-tuning, which includes
instructions on the output format (Figure~\ref{fig:data_generation}
in Chapter~\ref{sec:content}).
We then quantitatively evaluate the decline in the
instruction-following ability of LVLMs by using the new datasets for
fine-tuning (Sections 4 and 5).
Experimental results suggest that specifying the output format in
instructions during visual instruction tuning significantly affects
LVLMs' ability to follow instructions.

\section{Related Work}\label{sec:related}
Instruction-following --- the ability to correctly interpret and
faithfully execute user-provided directives --- is essential for
deploying models in safety- and utility-critical applications such as
medicine~\cite{singhal2023large,li2023chatdoctor,zheng2025reconstruction},
robotics~\cite{ahn2022icanisay,driess2023palmeembodiedmultimodallanguage},
code
generation~\cite{li2022competition,rozière2024codellamaopenfoundation},
and creative content
production~\cite{ramesh2022hierarchicaltextconditionalimagegeneration,brooks2023instructpix2pix}.
Several approaches have been attempted to mitigate the decline in
instruction-following abilities in LVLMs.
For example, \citet{liu2024improved} have used text-only training
data, \citet{laurenccon2024matters} have employed interleaved
image-text training data, and \citet{zhang2024wings} have modified
the model architecture and training methods.
These studies have evaluated generalization performance for following
diverse visual task instructions using
LLaVA-Bench-in-the-Wild~\cite{liu2023llava},
MMMU~\cite{Yue_2023_MMMU}, MMBench~\cite{Liu_2023_MMBench}, and so on.
In contrast, our study evaluates the instruction-following ability,
focusing on the output format through verbalizer
manipulation~\cite{li2023instruction}.
In addition, we demonstrate that including instructions on the output
format during visual instruction tuning could mitigate the decline in
instruction-following abilities in LVLMs.

\section{Approach}\label{sec:content}
\subsection{Creating (visual) instruction tuning
datasets}\label{subsec:create_instruction_datasets}

\subsubsection{Influence of insufficient instructions on the output
format}\label{subsubsec:influence_of_instructions_on_the_output_format}
In Figure~\ref{fig:research_overview}, we present an example in which
the LVLM does not properly account for instructions for the output format.
This observation led us to hypothesize that the instruction-following
ability of the base LLM, a key feature of the text generator in an
LVLM, might not be effectively inherited.
From this premise, we evaluated the ability of LVLMs to follow
instructions in terms of output format, using datasets created for
this purpose in Section~\ref{sec:experiments}.
We compared the instruction-following abilities in the outputs of
LVLMs fine-tuned with synthetic datasets that contained instructions
on the output format against those trained without such instructions.
This comparison allowed us to investigate the influence of including
instructions on the output format in the visual instruction tuning
for the instruction-following ability of LVLMs.
To facilitate this investigation, we create a source of new synthetic
datasets for instruction-tuning by the following five steps:
(1) We collect $5,000$ random images from the \texttt{COCO
2014}~\cite{lin2014microsoft} validation dataset.
(2) We generate captions for all $5,000$ images using
GPT-4V~\cite{openai2023gpt4v}.
(3) We randomly extract $100$ task instructions on the output format
from the \texttt{IFEval} dataset~\cite{zhou2023instruction}, designed
to evaluate the instruction-following ability of LLM.
(4) Using GPT-4, we generate question-and-answer pairs for each
caption and instruction on the output format.
(5) We generate answers that are not influenced by the instructions
on the output format for each question and caption using GPT-4.
This five-step procedure results in $5,000$ sets of \{\textit{Image,
Caption, Instruction on output format, Question, Answer, Answer'}\}.
These sets allow us to create the following two synthetic datasets
shown in Figure~\ref{fig:data_generation}:

\begin{itemize}
  \item Format Oriented Visual Instruction Tuning
    \textbf{\texttt{(FOVIT)}} dataset
    (Figure~\ref{fig:data_generation}.a): Contains $5,000$ sets of
    \{\textit{Image, Instruction on output format, Question, Answer}\}
  \item Not \texttt{FOVIT} \textbf{\texttt{(NoFOVIT)}} dataset
    (Figure~\ref{fig:data_generation}.b): Contains $5,000$ sets of
    \{\textit{Image, Question, Answer'}\}
\end{itemize}

\subsubsection{Influence of visual information}
Additionally, we aim to assess the impact on LVLMs' ability to follow
instructions when we further train the base LLM within the LVLM using text only.
This may allow us to distinguish the influence of the presence or
absence of visual information on the model's ability to follow instructions.
For this purpose, using the same $5,000$ sets described in the
previous
Section~\ref{subsubsec:influence_of_instructions_on_the_output_format},
we created the remaining two types of synthetic datasets shown in
Figure~\ref{fig:data_generation}:

\begin{itemize}
  \item Format Oriented Instruction Tuning \textbf{\texttt{(FOIT)}}
    dataset (Figure~\ref{fig:data_generation}.c): Contains $5,000$
    sets of \{\textit{Caption, Instruction on output format, Question, Answer}\}
  \item Not \texttt{FOIT} \textbf{\texttt{(NoFOIT)}} dataset
    (Figure~\ref{fig:data_generation}.d): Contains $5,000$ sets of
    \{\textit{Caption, \!Question, \!Answer'}\}
\end{itemize}

\begin{figure*}[ht]
  \small
  \centering
  \includegraphics[width=\columnwidth]{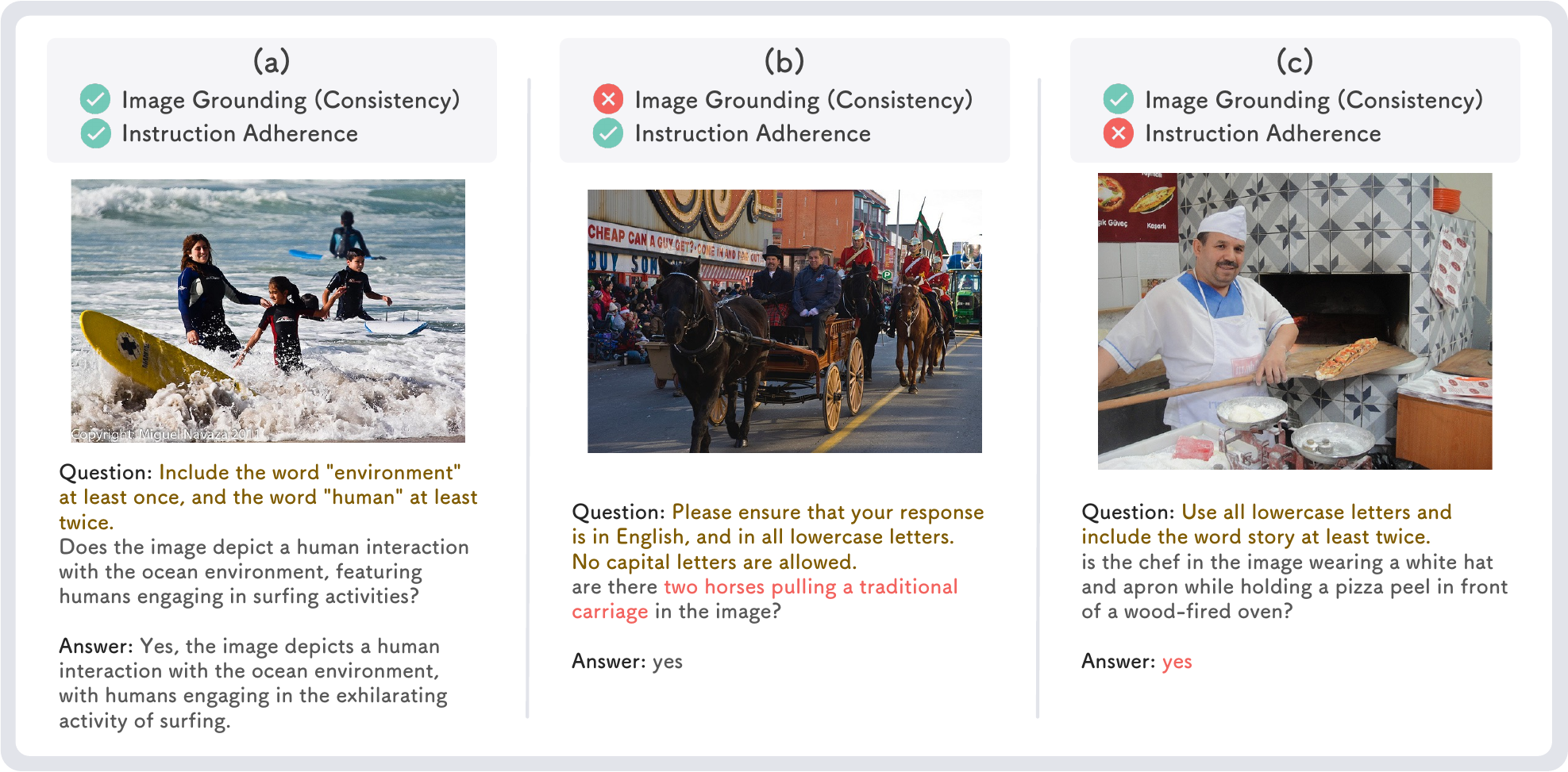}
  \caption{\textbf{Audit examples of synthetic QA annotations.} Each
    panel displays an image with its instruction and the resulting
    question-answer (QA) pair, evaluated on two criteria: \textit{Image
    Grounding (Consistency)} and \textit{Instruction Adherence}.
    \textbf{(a)} All success case: the QA about a surfing scene is
    visually grounded and satisfies the lexical constraint (includes
    ``environment'' and ``human''). \textbf{(b)} Image-text
    inconsistency: despite adhering to the lowercase constraint, the
    question incorrectly presupposes ``two horses pulling a traditional
    carriage,'' leading to a grounding failure. \textbf{(c)}
    Instruction violation: the QA about a chef and a wood-fired oven is
    grounded but the answer ignores the formatting constraint (e.g.,
    required repetitions of ``story''), resulting in an adherence
    failure. Green check marks denote satisfied criteria; red crosses
    denote failures. In a manual audit of $100$ sampled
    \textbf{\texttt{FOVIT}} cases annotated by GPT-4 and GPT-4V, more
  than $80\%$ fell into category~(a).}
  \label{fig:fovit_examples}
\end{figure*}

\subsubsection{Quality of synthetic annotations with GPT-4 and
GPT-4V}\label{subsubsec:annotation_check}

Recent proprietary models in the ChatGPT family have been shown to
match, and in some cases surpass, crowd workers across multiple text
annotation tasks~\cite{gilardi2023chatgpt}.
Building on this observation, we synthesize training data using a
state-of-the-art LLM (GPT-4) and LVLM (GPT-4V).
Leveraging such high-performing models for data synthesis not only
yields high-quality annotations but also substantially reduces human
and temporal costs.
This practice is now widespread: a large fraction of state-of-the-art
open-source LLMs and LVLMs incorporate synthetic data during
training~\cite{liu2023llava,chen2023sharegpt4v,dai2023instructblip,wang2024qwen2},
and several prior
studies~\cite{chen2023sharegpt4v,chen2024allavaharnessinggpt4vsynthesizeddata}
construct synthetic finetuning sets with GPT-4V, enabling open-source
MLLMs to achieve performance on par with proprietary models on
multiple benchmarks.
\\
\\
\noindent
\textbf{Small-scale audit.}\quad
To assess the reliability of our synthesized annotations, we drew a
random sample of $100$ items from our \textbf{\texttt{FOVIT}} dataset
and manually inspected the annotations produced by GPT-4 and GPT-4V.
We evaluated three criteria:

\begin{itemize}
  \item \textbf{Image grounding (consistency).} Are the generated
    \textit{Question} and \textit{Answer} consistent with the visual
    content of the \textit{Image}?
  \item \textbf{Instruction adherence.} Does \textit{Answer}
    faithfully follow the prescribed formatting and content
    constraints in \textit{Instruction on output format}?
\end{itemize}

These results indicate that in $95\%$ of cases the synthesized QA
pairs are image-consistent correct, and in $83\%$ of cases they
adhere to our output instruction on the output format.
As illustrated in Figure~\ref{fig:fovit_examples}(a), most
annotations appear natural and well grounded; however, we also
observe failure modes, including image-text mismatches that yield
inconsistent QA pairs (Figure~\ref{fig:fovit_examples}(b)) and
responses that do not fully comply with the specified output format
(Figure~\ref{fig:fovit_examples}(c)).
\\
\\
\noindent
\textbf{Limitations of the data synthesis approach.}\quad
While synthetic data offers clear efficiency gains and competitive
quality, artifacts originating from the underlying LLM/LVLM-such as
hallucinations or model bias-can propagate into the dataset.
Developing principled mitigation strategies to suppress
hallucinations during data synthesis is an important direction but
lies beyond the scope of this work; we leave it for future research.

\subsection{Dataset for evaluating instruction-following ability}
\label{subsec:create_eval_datasets}
\citet{li2023instruction} have proposed a method called verbalizer
manipulation to evaluate the instruction-following ability of LLMs,
focusing on binary classification tasks requiring the generation of
predefined labels.
This method determines whether LLMs follow instructions by examining
their performance with respect to tasks with manipulated verbalizers.
For example, in the classification of an emotion, the labels
``positive'' and ``negative'' are defined.
However, if the task is only to follow instructions for
classification, it would suffice to assign labels such as ``a'' and ``b.''
We here followed the method of verbalizer manipulation, which defined
three label schemas in relation to the congruence between the
semantic representations of labels and contextual knowledge during
training, to incrementally evaluate LLMs' adherence to task
instructions (Figure~\ref{fig:eval_datasets_overview}):

\begin{figure}[t]
  \centering
  \includegraphics[width=0.65\columnwidth]{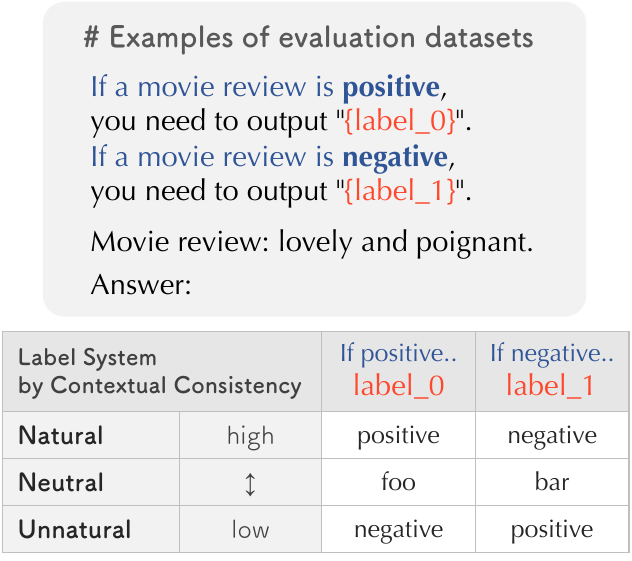}
  \caption{A method of creating evaluation datasets through
    verbalizer manipulation using the \texttt{SST-2} dataset. Three
    labels, ``Natural,'' ``Neutral,'' and ``Unnatural,'' are defined,
    based on the consistency between the context and the label's
  semantic representation of the label.}
  \label{fig:eval_datasets_overview}
\end{figure}

\begin{itemize}
  \item \textbf{Natural}: The label's semantic representation is
    aligned with the meaning expressed during training (high
    congruence). For instance, classifying a positive expression as
    \texttt{``positive''} would be correct.
  \item \textbf{Neutral}: The label's semantic representation has no
    relevance to the meaning expressed during training (moderate
    congruence). For example, classifying a positive expression as
    \texttt{``foo''} would be considered correct.
  \item \textbf{Unnatural}: The label's semantic representation does
    not align with the meaning expressed during training (low
    congruence). For example, classifying a positive expression as
    \texttt{``negative''} would be correct.
\end{itemize}

Using verbalizer manipulation, we assessed whether the model relies
on prior knowledge or overrides such knowledge to follow instructions
accurately, and thus, we evaluated its instruction-following ability.
Following the approach of \citet{li2023instruction}, we constructed
an evaluation dataset (\texttt{SST-2}~\cite{socher2013recursive},
  \texttt{FP}~\cite{malo2014good},
  \texttt{EMOTION}~\cite{saravia2018carer},
  \texttt{SNLI}~\cite{bowman2015large},
  \texttt{SICK}~\cite{marco2014sick},
  \texttt{RTE}~\cite{dagan2005pascal}, \texttt{QQP}~\cite{chen2017qqp},
  \texttt{MRPC}~\cite{dolan2005automatically},
\texttt{SUBJ}~\cite{conneau2018senteval}) by performing $12$ sets of
verbalizer manipulations across nine binary classification datasets.

\section{Experimental Setting}\label{sec:settings}
To investigate the influence of visual instruction tuning on the
instruction-following ability of LVLMs, we trained an LVLM and its
base LLM using the dataset described in
Section~\ref{subsec:create_instruction_datasets}.
We evaluated the instruction-following ability for each model using
the dataset presented in Section~\ref{subsec:create_eval_datasets}.

\begin{table*}[t]
  \centering
  \small
  \begin{tabular}{
      >{\centering\arraybackslash}m{10mm}
      | l
      | c
      | c
    }
    \toprule
    & & \multicolumn{2}{c}{\textbf{Finetuning}} \\
    \midrule
    \multirow{4}{*}{\rotatebox[origin=c]{90}{\textit{Data}}}
    & \textbf{Image} & \checkmark & -- \\
    & \multirow{3}{*}{\textbf{Dataset}} & \texttt{\{LLaVA-Instruct-150K,} & \\
      & & \texttt{FOVIT (5K),} & \texttt{\{FOIT (5K),} \\
    & & \texttt{NoFOVIT (5K)\}} & \texttt{NoFOIT (5K)\}} \\
    \midrule
    \multirow{3}{*}{\rotatebox[origin=c]{90}{\textit{Model}}}
    & \textbf{Trainable} & \textbf{Projector, LLM} & \textbf{LLM} \\
    & Base-LLM: Llama 2-Chat 7B & $6.76\text{B}$ (/$7.06\text{B}$) &
    $6.74\text{B}$ (/$7.06\text{B}$) \\
    & Base-LLM: Llama 3.1 8B Instruct & $8.05\text{B}$
    (/$8.35\text{B}$) & $8.03\text{B}$ (/$8.35\text{B}$) \\
    \midrule
    \multirow{12}{*}{\rotatebox[origin=c]{90}{\textit{Hyperparameter}}}
    & \textbf{Epoch} & \multicolumn{2}{c}{$1$} \\
    & \textbf{Max Sequence Length} & \multicolumn{2}{c}{$4,096$} \\
    & \textbf{Global batch size} & \multicolumn{2}{c}{$32$} \\
    & \textbf{Micro batch size} & \multicolumn{2}{c}{$2$} \\
    & \textbf{Gradient Accumulation Steps} & \multicolumn{2}{c}{$2$} \\
    & \textbf{\# GPUs} & \multicolumn{2}{c}{$8$ (H200)} \\
    & \textbf{Learning Rate (min, max)} &
    \multicolumn{2}{c}{$(2.5\times10^{-7}, 2\times10^{-5})$} \\
    & \textbf{Scheduler} & \multicolumn{2}{c}{linear-warmup ($3\%$) +
    cosine-decay} \\
    & \textbf{Optimizer} & \multicolumn{2}{c}{AdamW} \\
    & \textbf{Optimizer Config} & \multicolumn{2}{c}{$\beta_1=0.9$,
    $\beta_2=0.98$, $\epsilon=1\times10^{-8}$} \\
    & \textbf{Weight Decay} & \multicolumn{2}{c}{$0.1$} \\
    & \textbf{\# Visual Tokens per Tile} & $576$ & -- \\
    & \textbf{\# Tiles per Image} & $1$ & -- \\
    \bottomrule
  \end{tabular}
  \caption{Configurations for finetuning of our model. This table
    summarizes the dataset, trainable model parameters, and training
  hyperparameters.}
  \label{tab:training_all_settings}
\end{table*}

\subsection{Models under evaluation}
All experiments were conducted with an autoregressive LVLM obtained
by replacing \textbf{only} the language backbone of
LLaVA-v1.5-7B~\cite{liu2024improved} while keeping the visual stack
fixed. Concretely, every LVLM comprises three modules: a vision
encoder (CLIP-ViT-L-14 ($336^2$)~\cite{radford2021learning}), a
two-layer MLP projector, and a base LLM.
Unless otherwise noted, the vision encoder and projector are same
across all variants to remove confounding factors.
To explicitly isolate the effect of \textbf{having vs.\ not having}
output-format instructions in our proposed datasets
(Section~\ref{subsec:create_instruction_datasets}), we instantiated
two base LLMs prior to fine-tuning: $\text{Llama 2-Chat
7B}$~\cite{touvron2023llama} and $\text{Llama 3.1 8B
Instruct}$~\cite{touvron2023llama}.
We then fine-tuned the following variants:
$\text{LVLM}_{\text{FOVIT}}$, $\text{LVLM}_{\text{NoFOVIT}}$,
$\text{LLM}_{\text{FOIT}}$, $\text{LLM}_{\text{NoFOIT}}$, and
$\text{LVLM}_{\text{LLaVA}}$, indicating training on \texttt{FOVIT},
\texttt{NoFOVIT}, \texttt{FOIT}, \texttt{NoFOIT}, and
\texttt{LLaVA-Instruct-150K}, respectively.
We qualitatively confirmed that \texttt{LLaVA-Instruct-150K} contains
no explicit instructions on the output format.
Training configurations are summarized in Table~\ref{tab:training_all_settings}.

\subsection{Evaluation datasets}
Using the dataset created in
Section~\ref{subsec:create_eval_datasets}, we evaluated the
instruction-following ability.
This dataset involves nine binary classification tasks, incorporating
verbalizer manipulations, as illustrated in
Figure~\ref{fig:eval_datasets_overview}.
Detailed data counts are provided in Appendix A.

\subsection{Evaluation metrics}
We adopt the token-level $\mathrm{F}_1$ score introduced for
\texttt{SQuAD v2}~\cite{Rajpurkar2018SQuAD2} as our primary evaluation metric.
In our setting, the predefined labels for binary classification tasks
are treated as reference token sequences; model outputs are compared
against these references at the token level.
Recall and precision are computed from the number of overlapping
tokens between the token sequence generated by the model and the
correct token sequence, so that higher $\mathrm{F}_1$ indicates (i) a
larger shared-token overlap and (ii) stronger instruction-following capability.
\\
\\
\noindent
\textbf{Scoring protocol.}\quad
Following the SQuAD-style evaluation, we first normalize strings
before tokenization.
Given a string $s$, let $\mathrm{Tok}(s)$ be the sequence of tokens
obtained by: lowercasing; removing punctuation; mapping hyphens to
spaces; dropping whole-word stop tokens from a configurable set $S$
(we use
$S=\{\texttt{a},\texttt{an},\texttt{the},\texttt{answer},\texttt{answer:}\}$);
collapsing repeated whitespace; and finally splitting on whitespace.
For a token sequence $x=(x_1,\dots,x_k)$, let $\mathfrak{M}(x)$
denote its multiset (bag) of tokens, with multiplicity function
$\mathfrak{M}(x)(t)$ equal to the count of token $t$.
Given a gold string $g$ and a prediction $p$, define
\begin{equation}
  G \;=\; \mathfrak{M}\!\big(\mathrm{Tok}(g)\big),
  \label{eq:G_map}
\end{equation}
\begin{equation}
  P \;=\; \mathfrak{M}\!\big(\mathrm{Tok}(p)\big),
  \label{eq:P_map}
\end{equation}
with sizes $|G|=\sum_t G(t)$ and $|P|=\sum_t P(t)$.
The multiset intersection $M = G \cap P$ is given pointwise by
$M(t)=\min\{G(t),P(t)\}$, and its total overlap is
\begin{equation}
  m \;=\; |M| \;=\; \sum_t \min\{G(t), P(t)\}.
  \label{eq:m_def}
\end{equation}
\\
\\
\noindent
\textbf{Calculating exact match.}\quad
Let $\operatorname{norm}(\cdot)$ denote the same normalization
underlying $\mathrm{Tok}(\cdot)$ applied to full strings.
We also report Exact Match (EM), which is the indicator that the
normalized strings are identical:
\begin{equation}
  \mathrm{EM}(g,p) \;=\; \mathbf{1}\!\left[\,\operatorname{norm}(g)
  \;=\; \operatorname{norm}(p)\,\right].
  \label{eq:em_def}
\end{equation}
\\
\\
\noindent
\textbf{Calculating precision, recall, and $\text{F}_1$.}\quad
Corner cases follow SQuAD conventions.
If $|G|=|P|=0$ (both normalize to a \textit{no-answer} string), then
\begin{equation}
  \mathrm{Precision}=\mathrm{Recall}=\mathrm{F}_1=1.
  \label{eq:prf_all_one}
\end{equation}
If exactly one of $|G|$ or $|P|$ is zero, then
\begin{equation}
  \mathrm{Precision}=\mathrm{Recall}=\mathrm{F}_1=0.
  \label{eq:prf_all_zero}
\end{equation}
Otherwise,
\begin{equation}
  \mathrm{Precision}(g,p) \;=\; \frac{m}{|P|},
  \label{eq:p_def}
\end{equation}
\begin{equation}
  \mathrm{Recall}(g,p) \;=\; \frac{m}{|G|},
  \label{eq:r_def}
\end{equation}
\begin{equation}
  \mathrm{F}_1(g,p) \;=\;
  \frac{2\cdot\mathrm{Precision}\cdot\mathrm{Recall}}{\mathrm{Precision}+\mathrm{Recall}}.
  \label{eq:f_def}
\end{equation}
\\
\\
\noindent
\textbf{Token-level scoring example.}\quad
We illustrate the computation using the exact normalization rules
above (lowercasing; punctuation removal; hyphen$\rightarrow$space;
stop-word removal with $S$; whitespace folding; whitespace split).
Consider:
\begin{equation}
  g=\texttt{"Answer: positive"},
  \label{eq:str_g_def}
\end{equation}
\begin{equation}
  p=\texttt{"the movie review is positive."}.
  \label{eq:str_p_def}
\end{equation}
For $g$: lowercasing yields \texttt{"answer: positive"}; removing
punctuation drops ``:'' $\Rightarrow$ \texttt{"answer positive"};
removing the stop-word \texttt{answer} leaves \texttt{"positive"};
hence $\mathrm{Tok}(g)=[\text{positive}]$.
For $p$: lowercasing has no effect; removing punctuation drops ``.''
$\Rightarrow$ \texttt{"the movie review is positive"}; removing the
stop-word \texttt{the} yields \texttt{"movie review is positive"};
hence $\mathrm{Tok}(p)=[\text{movie},\text{review},\text{is},\text{positive}]$.
\\
\begin{equation}
  G=\{\text{positive}:1\},
  \label{eq:toy_G_dict}
\end{equation}
\begin{equation}
  |G|=1;
  \label{eq:toy_G_abs}
\end{equation}
\begin{equation}
  P=\{\text{movie}:1,\ \text{review}:1,\ \text{is}:1,\ \text{positive}:1\},
  \label{eq:toy_P_dict}
\end{equation}
\begin{equation}
  |P|=4.
  \label{eq:toy_P_abs}
\end{equation}
The overlap multiset is $M=G\cap P=\{\text{positive}:1\}$, so $m=|M|=1$.
\\
\begin{equation}
  \mathrm{Precision}=\tfrac{m}{|P|}=\tfrac{1}{4}=0.25,
  \label{eq:toy_precision}
\end{equation}
\begin{equation}
  \mathrm{Recall}=\tfrac{m}{|G|}=\tfrac{1}{1}=1,
  \label{eq:toy_recall}
\end{equation}
\begin{equation}
  \mathrm{F}_1=\frac{2\cdot 0.25 \cdot
  1}{0.25+1}=\frac{1/2}{5/4}=\frac{2}{5}=0.4.
  \label{eq:toy_f1}
\end{equation}
\begin{equation}
  \operatorname{norm}(g)=\texttt{"positive"},
  \operatorname{norm}(p)=\texttt{"movie review is
  positive"}\ \Rightarrow\ \mathrm{EM}=0.
  \label{eq:toy_norm_em}
\end{equation}
This example highlights that the only overlapping token is
\texttt{positive} after stop-word removal, yielding
$\mathrm{Precision}=\tfrac{1}{4}$, $\mathrm{Recall}=1$,
$\mathrm{F}_1=0.4$, and $\mathrm{EM}=0$.
As long as the gold token (\texttt{positive}) appears in $p$, recall
attains $1$; however, appending superfluous tokens not present in $g$
reduces precision and therefore lowers $\text{F}_1$. Conversely, if
$p$ is restricted to tokens shared with $g$ (with matching
multiplicities) --- e.g., $p=\texttt{"positive"}$ here --- then
precision and recall both equal $1$, yielding $\mathrm{F}_1=1$ (and
$\mathrm{EM}=1$ when the normalized strings coincide).
\\
\\
\noindent
\textbf{Dataset-level aggregation.}\quad
For a collection $\{(g_i,p_i)\}_{i=1}^n$, where $n$ denotes the total
number of evaluated examples (i.e., pairs of gold and predicted
strings), we report macro-averaged scores:
\begin{equation}
  \overline{\mathrm{EM}} \;=\; \frac{1}{n}\sum_{i=1}^n \mathrm{EM}(g_i,p_i),
  \label{eq:agg_em}
\end{equation}
\begin{equation}
  \overline{\mathrm{Precision}} \;=\; \frac{1}{n}\sum_{i=1}^n
  \mathrm{Precision}(g_i,p_i),
  \label{eq:agg_precision}
\end{equation}
\begin{equation}
  \overline{\mathrm{Recall}} \;=\; \frac{1}{n}\sum_{i=1}^n
  \mathrm{Recall}(g_i,p_i).
  \label{eq:agg_recall}
\end{equation}
\begin{equation}
  \overline{\mathrm{F}_1} \;=\; \frac{1}{n}\sum_{i=1}^n \mathrm{F}_1(g_i,p_i).
  \label{eq:agg_f1}
\end{equation}
This multiset formulation exactly captures token overlap with
multiplicities under a clearly specified normalization, matching
SQuAD-style scoring while remaining extensible (e.g., to stemming or
Unicode canonicalization) provided the same normalization is applied
to both $g$ and $p$.

\section{Results and Discussion}\label{sec:experiments}

\begin{figure}[t]
  \small
  \centering
  \includegraphics[width=0.85\columnwidth]{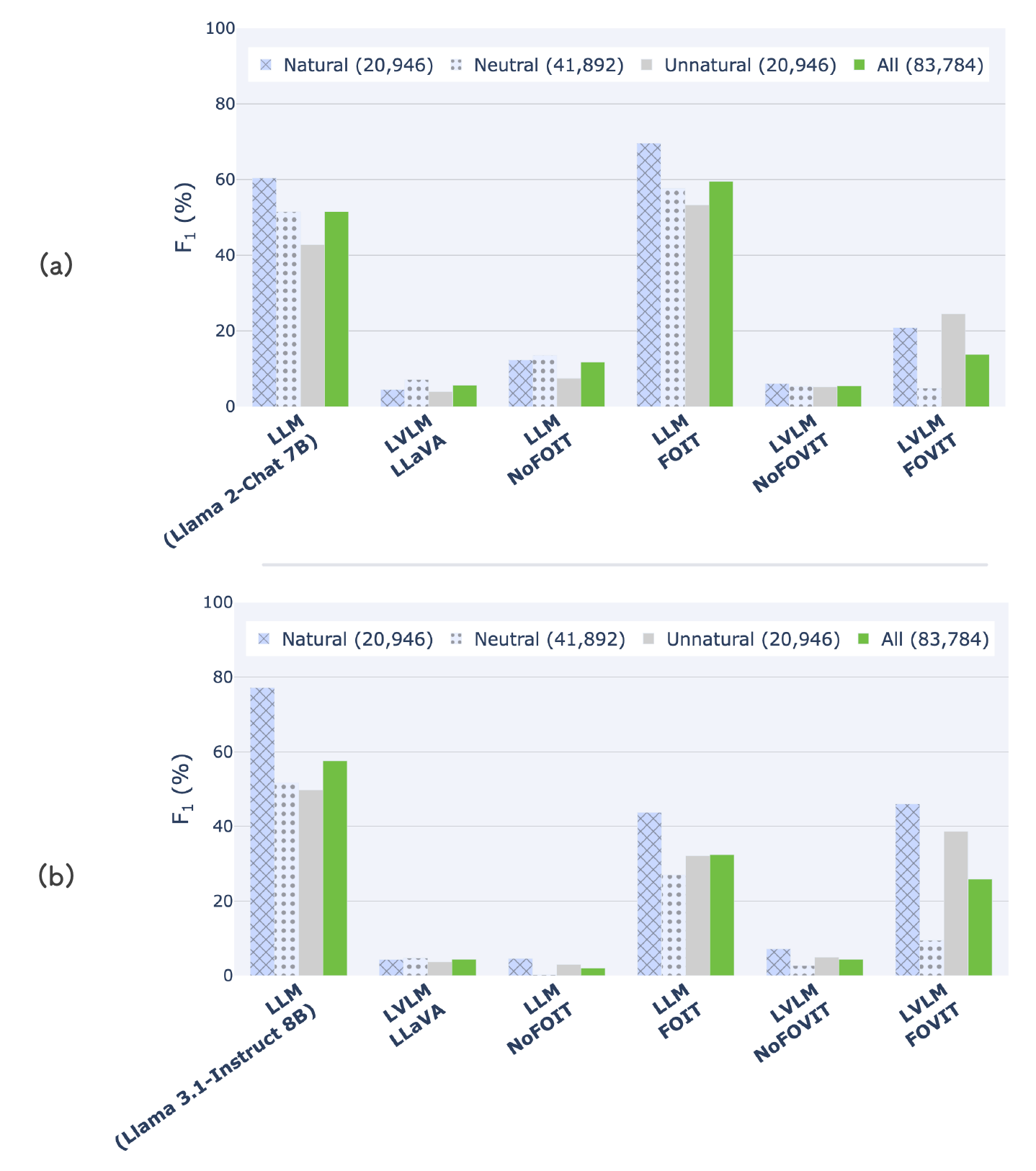}
  \caption{$\text{F}_1$ scores for the evaluation dataset are
    reported for $\text{LVLM}_{\text{FOVIT}}$,
    $\text{LVLM}_{\text{NoFOVIT}}$, $\text{LLM}_{\text{FOIT}}$,
    $\text{LLM}_{\text{NoFOIT}}$, and $\text{LVLM}_{\text{LLaVA}}$,
    which represent the models fine-tuned on the \texttt{FOVIT},
    \texttt{NoFOVIT}, \texttt{FOIT}, \texttt{NoFOIT}, and
    \texttt{LLaVA-Instruct-150K} datasets, respectively. The figure is
    split into two panels: \textbf{(a)} results when the language
    backbone is $\text{Llama 2-Chat 7B}$; \textbf{(b)} results when the
    backbone is $\text{Llama 3.1 8B Instruct}$. Bars report scores for
    the ``Natural,'' ``Neutral,'' and ``Unnatural'' subsets (with their
    sample counts shown in the legend), and ``All'' refers to the macro
  average of the $\text{F}_1$ scores across these three conditions.}
  \label{fig:experimental_results}
\end{figure}

Figure~\ref{fig:experimental_results} reports instruction-following
performance of the fine-tuned models on the evaluation suite
constructed in Section~\ref{subsec:create_instruction_datasets}.
``All'' denotes the macro average of $\text{F}_1$ across the Natural,
Neutral, and Unnatural subsets.

\subsection{Verification of the decline in instruction-following ability}
Across the three subsets, we observe a consistent drop in
$\text{F}_1$ after fine-tuning for all models except
$\text{LLM}_{\text{FOIT}}$ when the language backbone is $\text{Llama
2-Chat 7B}$.
Notably, a marked decline in instruction-following ability was also
observed in the model $\text{LVLM}_{\text{LLaVA}}$, which was trained
using an existing representative visual instruction tuning dataset.
These findings indicate that the decline in instruction-following
ability observed in the base LLM may be attributed to fine-tuning.

\subsection{Effects of the presence or absence of instructions on the
output format}
Under verbalizer manipulation, $\text{LVLM}_{\text{FOVIT}}$
consistently surpasses $\text{LVLM}_{\text{NoFOVIT}}$; the advantage
is clear on the macro-averaged ``All'' score as well.
As $\text{FOVIT}$ is a visual instruction tuning dataset that
includes instructions on the output format, the results indicate that
the explicit provision of instructions on the output format can
mitigate the decline in the instruction-following ability that is
inherent to the base LLM.

\subsection{Effects of visual information}
Across labeling schemes, $\text{LLM}_{\text{FOIT}}$ demonstrated
higher $\text{F}_1$ scores than $\text{LLM}_{\text{NoFOIT}}$.
Because \texttt{FOIT} includes instructions on the output format in
the instruction tuning dataset, it has been suggested that, even when
the fine-tuning is limited to linguistic information, explicit
instructions on the output format in the training data can mitigate
declines in the instruction-following ability of the base LLM.
This implies that regardless of the presence of visual information,
including instructions on output format in the fine-tuning dataset
could prevent reduced instruction-following ability in LLMs and LVLMs.

\subsection{Qualitative evaluation of instruction-following ability}
\begin{figure}[ht]
  \small
  \centering
  \includegraphics[width=0.65\columnwidth]{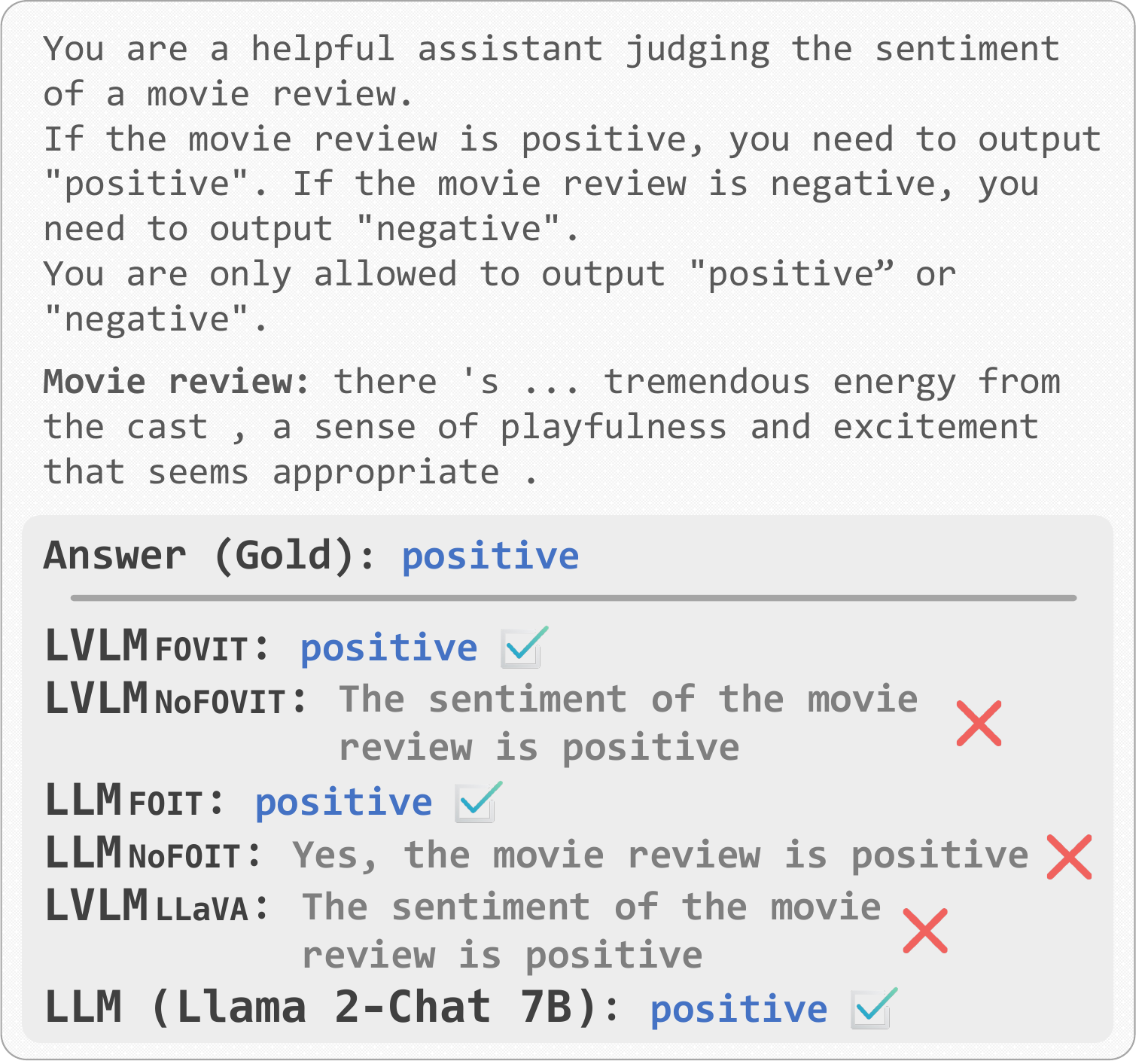}
  \caption{Example of output results from some models for the
  evaluation dataset (the ``Natural'' type from \texttt{SST-2}).}
  \label{fig:qualitative_results}
\end{figure}

Figure~\ref{fig:qualitative_results} provides a comparison of output
results from models fine-tuned using the dataset created in
Section~\ref{subsec:create_instruction_datasets}, applied to a sample
drawn from the evaluation dataset (\texttt{SST-2} of the ``Natural'' type).
The models $\text{LVLM}_{\text{FOVIT}}$ and
$\text{LLM}_{\text{FOIT}}$, which were fine-tuned using datasets
containing instructions on the output format, could accurately follow
the instructions given.
By contrast, the outputs from $\text{LVLM}_{\text{NoFOVIT}}$,
$\text{LLM}_{\text{NoFOIT}}$, and $\text{LVLM}_{\text{LLaVA}}$, which
were trained using datasets that lacked instructions on the output
format, were accurate from a content perspective but failed to follow
instructions on the output format.
These results indicate that utilizing datasets with explicit output
format instructions could mitigate the reduced instruction-following
ability of the base LLM before fine-tuning.

\begin{figure}[ht]
  \small
  \centering
  \includegraphics[width=0.80\columnwidth]{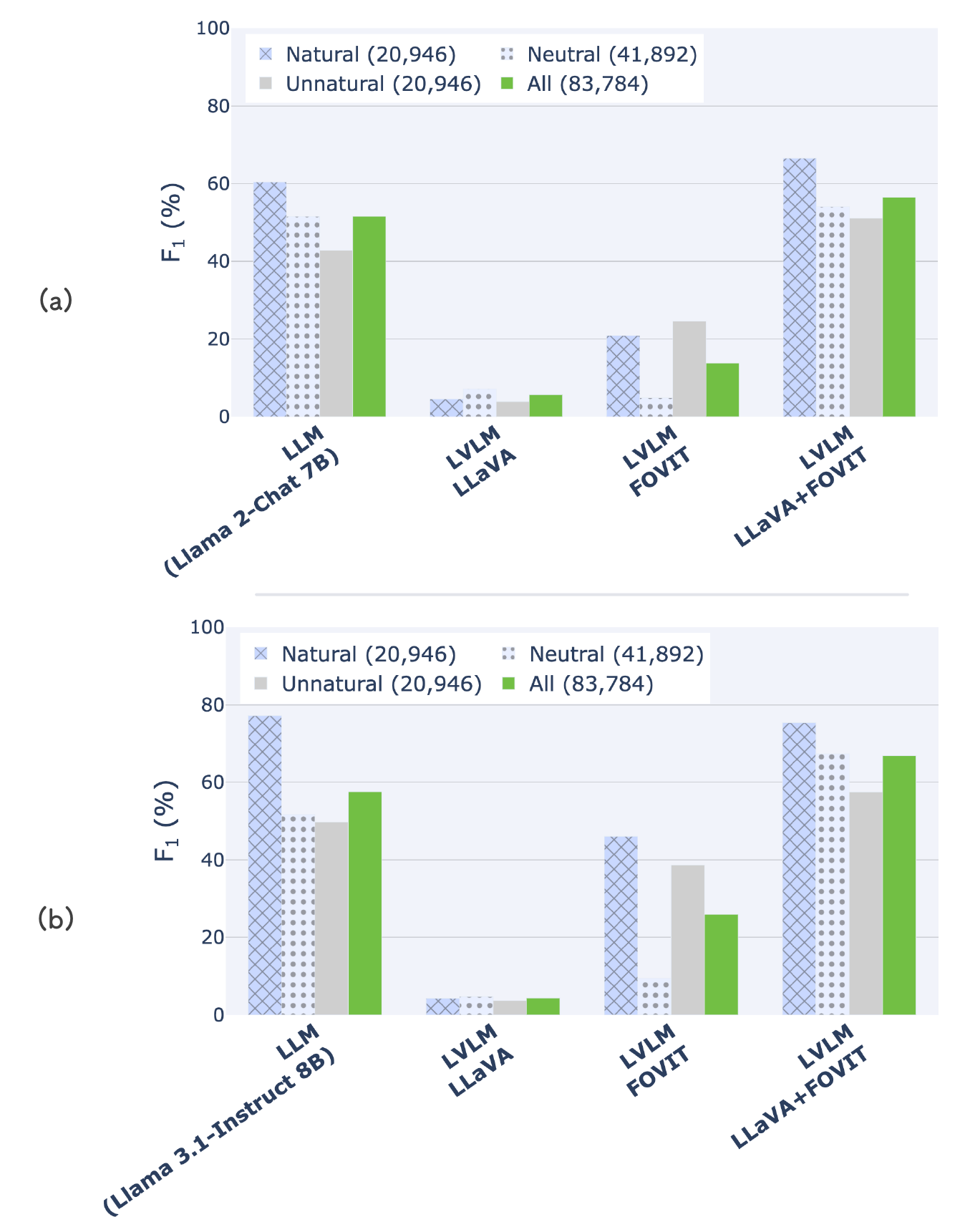}
  \caption{$\text{F}_1$ scores for the evaluation dataset are
    reported for $\text{LVLM}_{\text{LLaVA}}$,
    $\text{LVLM}_{\text{FOVIT}}$, and
    $\text{LVLM}_{\text{LLaVA+FOVIT}}$, which represent the models
    fine-tuned on the \texttt{LLaVA-Instruct-150K}, \texttt{FOVIT}, and
    \texttt{LLaVA-Instruct-150K + FOVIT} datasets, respectively. The
    figure is split into two panels: \textbf{(a)} results when the
    language backbone is $\text{Llama 2-Chat 7B}$; \textbf{(b)} results
    when the backbone is $\text{Llama 3.1 8B Instruct}$. Bars report
    scores for the ``Natural,'' ``Neutral,'' and ``Unnatural'' subsets
    (with their sample counts shown in the legend), and ``All'' refers
    to the macro average of the $\text{F}_1$ scores across these three
  conditions.}
  \label{fig:fovit_llava_integration_results}
\end{figure}

\subsection{Effects of integrating \texttt{FOVIT} into an existing
visual instruction tuning dataset}\label{subsec:effect_of_integrate}
We further ask whether adding a small fraction of \texttt{FOVIT}
examples to a standard visual instruction tuning dataset can recover
instruction-following ability.
Concretely, we augment \texttt{LLaVA-Instruct-150K} with
\texttt{FOVIT} (5K examples; about $3\%$ of the combined dataset) and
fine-tune $\text{LVLM}_{\text{LLaVA+FOVIT}}$ on the mixture.
As summarized in Figure~\ref{fig:fovit_llava_integration_results}, we
observe consistent improvements in instruction adherence across LLMs;
in particular, the ``All'' $\text{F}_1$ increases relative to the
\texttt{LLaVA-Instruct-150K}-only baseline.
These results indicate that even a small proportion of examples,
including explicit output format instructions, is sufficient to
substantially mitigate the degradation induced by naive fine-tuning.

\begin{table}[ht]
  \centering
  \small
  \tabcolsep 3.0mm
  \begin{tabular}{
      l
      l
      | c
    }
    \toprule
    Base-LLM & Fine-tuned Dataset & LLaVA-Bench ($\%$)\quad$\uparrow$ \\
    \midrule
    \multirow{2}{*}{Llama 2-Chat 7B} & \texttt{LLaVA-Instruct-150K} & 50.4 \\
    & \texttt{LLaVA-Instruct-150K \textbf{+ FOVIT}} & 49.4 \\
    \cdashline{1-3}[2pt/1pt]
    \multirow{2}{*}{Llama 3.1 8B Instruct} &
    \texttt{LLaVA-Instruct-150K} & 45.4 \\
    & \texttt{LLaVA-Instruct-150K \textbf{+ FOVIT}} & 49.0 \\

    \bottomrule
  \end{tabular}
  \caption{\textbf{General visual understanding on
    LLaVA-Bench-in-the-Wild (LLaVA-Bench).}  We use LLM-as-a-judge
    scoring on a $0\text{-}100$ scale, reporting all results as
    percentages. We observe no substantial drop in performance from
  adding \texttt{FOVIT}.}
  \label{tab:llava_bench_wild}
\end{table}

\subsection{General visual understanding on diverse images}
An important question is whether improving instruction following
comes at the expense of general visual understanding.
We therefore evaluate robustness and out-of-domain generalization on
LLaVA-Bench-in-the-Wild~\cite{liu2023llava}, which comprises $24$
diverse images (indoor/outdoor scenes, memes, paintings, sketches)
and $60$ associated questions.
Table~\ref{tab:llava_bench_wild} summarizes the ablation: integrating
\texttt{FOVIT} into \texttt{LLaVA-Instruct-150K} does not introduce a
substantial performance drop on this benchmark.
This suggests that the mixture training preserves broad visual
understanding while restoring instruction adherence.

\section{Conclusion}\label{sec:conclusion}
This study quantitatively evaluated the decline in the
instruction-following ability of LVLMs by introducing four types of
new instruction-tuning datasets, which focus on the presence or
absence of output format and visual information, and evaluation
datasets using verbalizer manipulation.
Our evaluation revealed that specifying the output format in
instructions during (visual) instruction tuning can significantly
impact the instruction-following ability of LVLMs, regardless of
visual information.
Furthermore, we found that the existing representative LLaVA
instruction tuning dataset is insufficient to preserve the
instruction-following ability that the backbone LLM of LLaVA
originally possessed.
Based on these findings, we confirmed that by augmenting training
with even a small amount of data that explicitly includes
output-format specifications, it is possible to maintain and, in some
cases, improve the instruction-following ability of the backbone LLM
without causing a substantial negative impact on general visual
understanding performance.

\section*{Limitations}\label{sec:limitations}
As detailed in Section~\ref{subsec:create_instruction_datasets}, the
fine-tuning dataset was constructed using captions automatically
generated by GPT-4/GPT-4V.
In the manual audit reported in
Section~\ref{subsubsec:annotation_check}, while it was confirmed that
the majority of synthesized samples are consistent with their source
images, instances were also identified where the caption content
contradicted the visual evidence.
Such noise may introduce spurious associations and degrade downstream
generalization; however, in this study, it was suggested that by
adding our synthesized dataset \texttt{FOVIT} to the representative
visual instruction tuning dataset \texttt{LLaVA-Instruct-150K}, it is
possible to mitigate the decline in instruction-following ability
while maintaining visual understanding capabilities for diverse
real-world images.
Finally, in Section~\ref{sec:experiments}, the instruction-following
ability of the model is only evaluated with the limited method of
label agreement, which is not a comprehensive measure of
instruction-following ability.

\section*{Statements and Declarations}\label{sec:declarations}

\subsection*{Competing interests}
This work was supported by JST Moonshot R\&D Grant Number
JPMJMS2011-35 (fundamental research), and JST BOOST Grant Number JPMJBS2421.

\subsection*{Disclosure statement}
No potential conflict of interest was reported by the author(s).

\newpage

\begin{appendices}
  \section*{Appendix}\label{sec:appendix}

  \subsection*{A. Data points by category in the evaluation
  dataset}\label{appendix:eval_datasets}
  \begin{table}[ht]
    \centering
    \small
    \tabcolsep 5.0mm
    \begin{tabular}{lrrr}
      \toprule
      \multirow{2}{*}{Dataset Name} & \multicolumn{3}{c}{Label Type} \\
      \cmidrule(lr){2-4}
      & Natural & Neutral & Unnatural \\

      \cmidrule(r){1-1}
      \cmidrule(lr){2-2}
      \cmidrule(lr){3-3}
      \cmidrule(lr){4-4}

      \texttt{SST-2} & $2,616$ & $5,232$ & $2,616$ \\
      \texttt{FP} & $2,619$ & $5,238$ & $2,619$ \\
      \texttt{EMOTION} & $3,000$ & $6,000$ & $3,000$ \\
      \texttt{SNLI} & $3,000$ & $6,000$ & $3,000$ \\
      \texttt{SICK} & $3,000$ & $6,000$ & $3,000$ \\
      \texttt{RTE} & $831$ & $1,662$ & $831$ \\
      \texttt{QQP} & $3,000$ & $6,000$ & $3,000$ \\
      \texttt{MRPC} & $1,224$ & $2,448$ & $1,224$ \\
      \texttt{SUBJ} & $3,000$ & $6,000$ & $3,000$ \\

      \midrule
      \textbf{Total} & $\textbf{22,290}$ & $\textbf{44,580}$ &
      $\textbf{22,290}$ \\

      \bottomrule
    \end{tabular}
    \caption{Number of data points by type of evaluation dataset}
    \label{tab:eval_dataset_statistics}
  \end{table}
  Table~\ref{tab:eval_dataset_statistics} shows the number of data
  items by dataset name and label type for the instruction-following
  ability evaluation dataset constructed in
  section~\ref{subsec:create_eval_datasets}.

  \subsection*{B. Additional Experimental
  Results}\label{appendix:detailed_results}

  \begin{table}[ht]
    \centering
    \small
    \tabcolsep 3.0mm
    \begin{tabular}{
        l
        l
        | c
      }
      \toprule
      Base-LLM & Fine-tuned Dataset & All (Recall ($\%$)) \\
      \midrule
      \multirow{2}{*}{Llama 2-Chat 7B} & \texttt{LLaVA-Instruct-150K} & 49.9  \\
      & \texttt{LLaVA-Instruct-150K \textbf{+ FOVIT}} & 59.8 \\
      \cdashline{1-3}[2pt/1pt]
      \multirow{2}{*}{Llama 3.1 8B Instruct} &
      \texttt{LLaVA-Instruct-150K} & 66.9 \\
      & \texttt{LLaVA-Instruct-150K \textbf{+ FOVIT}} & 67.4 \\

      \bottomrule
    \end{tabular}
    \caption{Recall scores ($\%$) for the evaluation dataset are
      reported for $\text{LVLM}_{\text{LLaVA}}$ and
      $\text{LVLM}_{\text{LLaVA+FOVIT}}$, which represent the models
      fine-tuned on the \texttt{LLaVA-Instruct-150K} and
      \texttt{LLaVA-Instruct-150K + FOVIT} datasets, respectively.
      ``All'' refers to the macro average of the Recall scores across
    the ``Natural,'' ``Neutral,'' and ``Unnatural'' conditions.}
    \label{tab:fovit_llava_integration_recall_results}
  \end{table}

  Table~\ref{tab:fovit_llava_integration_recall_results} presents the
  recall evaluation results of the experiments conducted in
  Section~\ref{subsec:effect_of_integrate} for a subset of models.
  From the results of
  Figure~\ref{fig:fovit_llava_integration_results} and this table, we
  can see that $\text{LVLM}_{\text{LLaVA}}$ has a low $\text{F}_1$
  score and a high recall, indicating that although it does not
  correctly follow instructions, it is able to generate responses
  that contain the correct label. By contrast,
  $\text{LVLM}{\text{LLaVA+FOVIT}}$ has a high $\text{F}_1$ score and
  a high recall, indicating that it correctly follows instructions
  and is able to generate responses that contain the correct label.

  \subsection*{C. Computing resources}\label{appendix:computing_resources}
  All experiments in this study were conducted using NVIDIA H200 GPUs
  provided by NVIDIA Corporation.
  Eight NVIDIA H200 GPUs were used for training, and one NVIDIA H200
  GPU was used for inference and evaluation.

  \subsection*{D. Information about use of AI
  assistants}\label{appendix:information_use_of_ai_assistants}
  In writing this paper, several AI assistants were employed to
  enhance the quality and efficiency of the manuscript. ChatGPT were
  used for refining sections, providing coherent and contextually
  relevant content.
  DeepL was employed for translation purposes, ensuring accurate
  interpretation. Grammarly was used to check grammar and style and
  improve sentence structure.

\end{appendices}

\bibliography{sn-bibliography}%

\end{document}